\documentclass[letterpaper]{article} 
\usepackage{aaai2027}  
\usepackage[hyphens]{url}  
\usepackage{graphicx} 
\usepackage{natbib}  
\usepackage{caption} 
\usepackage{amsmath,amssymb,amsfonts}
\usepackage{multirow}
\usepackage{booktabs}
\usepackage{textcomp}
\usepackage{pifont}
\usepackage[table]{xcolor}

\title{FedOGL: Combating Catastrophic Forgetting in Federated Open-World Multimodal Graph Learning}

\author{Zekai Chen, Haodong Lu, Shihao Li, Weiwei Ji, Xunkai Li, Xun Wu, Yinlin Zhu, Rong-Hua Li}
\affiliations{Beijing Institute of Technology\\
Beijing, China}

\begin{document}

\maketitle

\begin{abstract}
Federated graph learning enables collaborative training over decentralized graph data without sharing raw graph information. We study federated open-world multimodal graph learning, motivated by regional video platforms that maintain private interaction graphs over users, videos, comments, and sharing behaviors. A scammer may splice an authentic interview clip of a financial expert with unrelated promotional audio and contact details. Each component appears legitimate in isolation, but together they create the false impression that the expert endorses the investment scheme. As such risks evolve, clients must learn emerging classes from private multimodal graph streams, retain historical categories, and reject samples outside the known class space. In this setting, clients must learn emerging classes from private multimodal graph streams while preserving historical categories and rejecting samples outside the current known class space. The core challenge is catastrophic forgetting, which in federated multimodal graphs is not merely a classifier-level failure: old knowledge can be erased through modality-semantic overwriting, topology-induced structural erosion, and federated memory fragmentation. 
To address this challenge, we propose \textbf{FedOGL}, a semantic-structural memory preservation framework. On the client side, FedOGL preserves historical decision behavior through replay and task-start distillation, while protecting graph-propagation memory via projection onto a globally shared structure basis. On the server side, FedOGL maintains and transfers compact category prototypes to facilitate cross-client knowledge sharing without exposing raw graph data. Extensive experiments demonstrate that, compared with the best-performing baselines, FedOGL reduces performance degradation caused by catastrophic forgetting by \textbf{42.67\%}, while maintaining or improving performance on downstream tasks.

\end{abstract}

\section{Introduction}
\label{sec:introduction}

\begin{figure}[t]
    \centering
    \includegraphics[width=\linewidth]{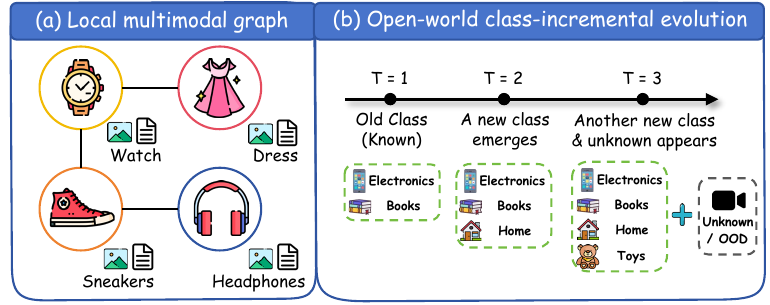}
    \caption{Federated open-world multimodal graph learning}
    \label{fig:intro_setting}
    \vspace{-10pt}
\end{figure}

Multimodal-attributed graphs have become a fundamental representation for real-world applications such as e-commerce search~\cite{wang2023fashionklip} and social media recommendation~\cite{wei2019mmgcn}, where graph topology captures high-order relational dependencies and multimodal attributes provide rich node-side semantics. In practice, such graphs are often distributed across organizations, platforms, or devices, making it infeasible to share raw nodes, edges, and multimodal contents due to privacy, ownership, and communication constraints. Federated learning (FL) enables collaborative training without raw-data exchange~\cite{mcmahan2017fedavg}, while federated graph learning (FGL) extends this paradigm to decentralized graph-structured data~\cite{he2021fedgraphnn,li2025openfgl}, providing a natural basis for multimodal FGL. Yet multimodal graph systems are particularly exposed to open-world evolution: new categories may emerge through unseen visual patterns, textual semantics, or cross-modal combinations, and different clients may encounter them asynchronously. A fixed and globally synchronized label space is therefore unrealistic, as clients must continually learn emerging categories, preserve historical graph knowledge, and reject samples outside the current known class space. This gives rise to federated open-world multimodal graph learning under class-incremental evolution, as illustrated in Fig.~\ref{fig:intro_setting}.

Existing methods address only parts of this setting. Multimodal FL/FGL methods enable decentralized representation sharing and coordinate modality semantics with graph structure~\cite{fedmvp2023,stage2026,prism2026}, but do not preserve historical semantic-structural memory under an evolving known/unknown class space. FGL methods such as FedSSP and FedIIH address structural or client heterogeneity~\cite{fedssp2024,fediih2025}, yet primarily assume fixed categories and provide limited support for evolving multimodal semantics. Open-world methods separately target continual retention with POWER~\cite{zhu2024federated}, graph OOD detection with TopoOOD and GRASP~\cite{li2024topoood,liu2023grasp}, or multimodal unknown recognition with CLIPN~\cite{clipn2023}; none jointly preserves multimodal semantics and graph structure across distributed clients as categories evolve. In this paper, catastrophic forgetting refers to the degradation of previously learned categories after the model is updated with emerging classes~\cite{rebuffi2017icarl}. Under open-world class expansion, such forgetting further distorts the known-class boundary, making unknown samples more likely to be over-accepted as known.

We argue that catastrophic forgetting in this new scenario is not merely a classifier-level failure, but a semantic-structural memory failure. Unlike standard continual learning, where forgetting is often treated as the loss of classifier-level decision regions, old knowledge in multimodal graphs is encoded through multiple coupled supports: modality-specific semantics, topology-dependent propagation patterns, and distributed category evidence. As illustrated in Sec.~\ref{sec:empirical}, once new tasks arrive, these supports can be damaged in three ways. \ding{182}\textbf{Modality-semantic overwriting}: new classes reshape modality encoders and overwrite the text-image semantics learned for old categories. \ding{183}\textbf{Topology-induced structural erosion}: historical categories rely on graph propagation directions, which can be disturbed by new-task optimization even when label-level knowledge is partially replayed. \ding{184}\textbf{Federated memory fragmentation}: old-class evidence is split into client-specific semantic-structural fragments under partial classes, neighborhoods, and modalities, while parameter averaging cannot explicitly identify and consolidate corresponding category memories. These coupled failures explain why historical categories are forgotten and why the evolving known/OOD boundary becomes unstable.

To address these challenges, we propose \textbf{FedOGL}, a semantic-structural memory preservation framework for federated open-world multimodal graph learning. Centered on anti-forgetting, FedOGL is designed around the above three memory failures. To counter modality-semantic overwriting, it preserves historical decision behavior with replay and task-start distillation. To counter topology-induced structural erosion, it constructs a global structure basis and projects structural gradients away from protected graph-propagation directions. To counter federated memory fragmentation, it converts local semantic-structural fragments into class-indexed prototypes, matches and merges corresponding memories across clients, and refines the global model over the resulting prototype graph. These designs enable FedOGL to preserve old-category semantic-structural memory while stabilizing the evolving known/OOD boundary under federated multimodal graph evolution.

\textbf{Our Contributions.}
\underline{\textbf{\emph{(1) New Perspective.}}} We formulate federated open-world multimodal graph learning under class-incremental evolution and reveal catastrophic forgetting in this setting as a semantic-structural memory failure caused by modality-semantic overwriting, topology-induced structural erosion, and federated memory fragmentation.
\underline{\textbf{\emph{(2) New Framework.}}} We propose FedOGL, a unified framework that combines replay and distillation, structure-basis protection, and prototype-based cross-client transfer to preserve semantic-structural memory under federated multimodal graph evolution.
\underline{\textbf{\emph{(3) SOTA Performance.}}} Across 6 datasets and 2 downstream tasks, FedOGL reduces catastrophic forgetting, measured by FM, by \textbf{42.67\%} and FPR$_{95}$ by \textbf{28.31\%} relative to the best-performing baseline, while maintaining or improving downstream-task performance.

\section{Preliminaries \& Problem Formulation}
\label{sec:problem}

We consider $K$ clients and a server over $T$ class-incremental stages. At stage $t$, client $k$ receives a private graph snapshot $\mathcal{G}_{k}^{t}=(\mathcal{V}_{k}^{t},\mathcal{E}_{k}^{t},\{\mathbf{X}_{k,m}^{t}\}_{m\in\mathcal M},\mathbf{y}_{k}^{t})$, where features may be unavailable for some modalities and $\mathbf{y}_{k}^{t}$ contains labels only for locally supervised nodes. Client $k$ observes a subset $\Delta\mathcal C_k^t$ of the new global classes $\Delta\mathcal C^t=\bigcup_k\Delta\mathcal C_k^t$; old examples are accessible only through its bounded replay memory.

Let $\mathcal C$ be the evaluation class universe. After learning stage $t$, the known and unknown spaces in Eq.~\eqref{eq:open_world_space} are
\begin{equation}
\begin{aligned}
\mathcal C_{\mathrm K}^{t}&=\bigcup_{\tau=1}^{t}\Delta\mathcal C^\tau,
\mathcal C_{\mathrm U}^{t}=\mathcal C\setminus\mathcal C_{\mathrm K}^{t},
f_{\theta^t}(v,\mathcal G_k^t)\in\mathcal C_{\mathrm K}^{t}\cup\{\mathrm{OOD}\}.&
\end{aligned}
\label{eq:open_world_space}
\end{equation}
Nodes from $\mathcal C_{\mathrm U}^{t}$ appear only in stage-$t$ evaluation; their semantic labels are hidden from training and become available only if their classes are introduced later. For node classification, $f_{\theta^t}$ predicts a known label or OOD. For cross-modal retrieval, modality-specific encoders rank the paired item among candidates from $\mathcal C_{\mathrm K}^{t}$ and reject queries from $\mathcal C_{\mathrm U}^{t}$. Clients transmit model updates and compact class-indexed statistics; nodes, modality records, and labels remain local.

\section{Related Work}
\label{sec:related}


\textbf{Multimodal FL/FGL.}
Multimodal FL/FGL further coordinates distributed modality semantics and graph structure. FedMVP federates visual prompt and prototype learning~\cite{fedmvp2023}. For graph-structured multimodal data, STAGE aligns cross-client semantics and regulates their structural propagation, while PRISM retrieves cross-client semantics and performs topology-aware imputation for modality-deficient clients~\cite{stage2026,prism2026}. MM-OpenFGL further formalizes multimodal FGL under joint modality, topology, and label heterogeneity~\cite{mmopenfgl2026}. However, these methods primarily optimize a fixed task and label space, without preserving old modality semantics and propagation directions during class expansion or consolidating class memories fragmented across clients.

\textbf{Open-world and continual graph learning.}
Open-world learning requires models to reject unseen classes~\cite{hendrycks2017baseline}, while continual graph learning studies sequential graph tasks and topology-dependent distribution shifts~\cite{ssrm2023}. POWER preserves knowledge on decentralized continual graphs~\cite{zhu2024federated}; TopoOOD and GRASP exploit graph structure for OOD detection~\cite{li2024topoood,liu2023grasp}; and CLIPN uses multimodal negative prompts to reject unknown concepts~\cite{clipn2023}. These methods isolate continual retention, structural OOD detection, or semantic unknown recognition. None jointly prevents modality-semantic overwriting, topology-induced structural erosion, and federated memory fragmentation. FedOGL unifies these concerns through semantic-structural memory preservation while stabilizing the evolving known/OOD boundary.

\section{Empirical Investigation}
\label{sec:empirical}

We diagnose three sources of forgetting in federated open-world multimodal graphs: modality-semantic overwriting, topology-induced structural erosion, and federated memory fragmentation.

\subsection{Modality-Semantic Overwriting}
\label{sec:empirical_semantic}

\begin{figure}[h]
    \centering
    \includegraphics[width=\linewidth]{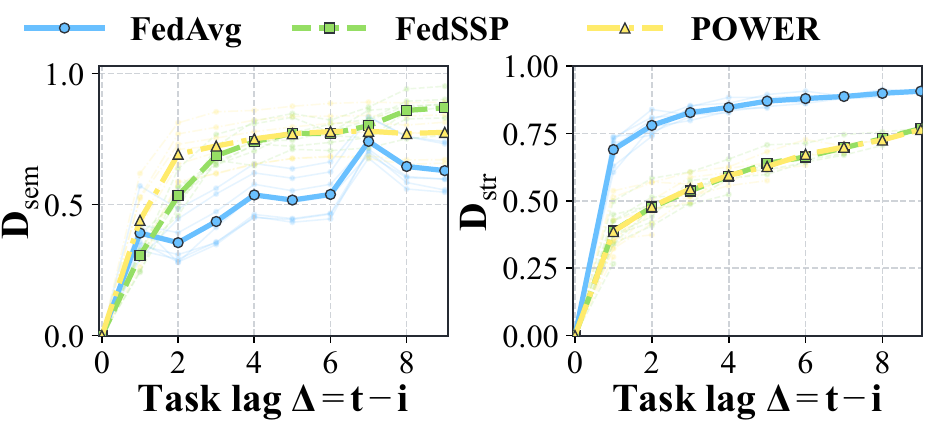}
    \caption{Semantic (left) and structural (right) memory drift over task lag. Thin curves denote classes or tasks; bold curves denote method averages.}
    \label{fig:semantic_structural_drift}
\end{figure}

On Grocery and Ele-Fashion, we measure the drift between an old class centroid when it first appears and after later task updates:
\begin{equation}
D_{\mathrm{sem}}(t)
=
\frac{1}{|\mathcal{C}_{\mathrm{old}}^{t}|}
\sum_{c\in\mathcal{C}_{\mathrm{old}}^{t}}
\left[
1-
\cos
\left(
\boldsymbol{\mu}_{c}^{\mathrm{sem},t_c},
\boldsymbol{\mu}_{c}^{\mathrm{sem},t}
\right)
\right].
\label{eq:semantic_drift}
\end{equation}

The left panel of Fig.~\ref{fig:semantic_structural_drift} shows increasing and dispersed semantic drift (Eq.~\eqref{eq:semantic_drift}) for all baselines as the task lag grows. FedSSP and POWER recover some old-class accuracy over FedAvg, yet their representations still move, indicating that predictive recovery alone does not ensure semantic stability.

\subsection{Topology-Induced Structural Erosion}
\label{sec:empirical_structural}

For historical task $i$, we compare the top-$r$ structural bases immediately after task $i$ and after task $t$ using Eq.~\eqref{eq:structural_drift}:
\begin{equation}
D_{\mathrm{str}}(i \!\rightarrow\! t)
=
\frac{1}{\sqrt{2r}}
\left\|
\mathbf{U}_{i}^{(i)}\mathbf{U}_{i}^{(i)\top}
-
\mathbf{U}_{i}^{(t)}\mathbf{U}_{i}^{(t)\top}
\right\|_{F}.
\label{eq:structural_drift}
\end{equation}
where zero indicates a preserved subspace and larger values indicate stronger erosion.

The right panel of Fig.~\ref{fig:semantic_structural_drift} shows steadily increasing drift in the ten-task Grocery stream. FedSSP and POWER improve over FedAvg, but still reach about $0.77$ after nine later tasks; existing regularization therefore slows rather than prevents topology-induced structural erosion.

\subsection{Federated Memory Fragmentation}
\label{sec:empirical_fragmentation}

We measure cross-client inconsistency by the mean cosine distance in Eq.~\eqref{eq:federated_fragmentation} between each local prototype $\mathbf{p}_{k,c}$ and its aggregated prototype $\bar{\mathbf{p}}_{c}$:
\begin{equation}
D_{\mathrm{fed}}
=
\frac{1}{|\mathcal{C}_{\mathrm{K}}|}
\sum_{c\in\mathcal{C}_{\mathrm{K}}}
\frac{1}{|\mathcal{K}_{c}|}
\sum_{k\in\mathcal{K}_{c}}
\left[
1-\cos(\mathbf{p}_{k,c},\bar{\mathbf{p}}_{c})
\right],
\label{eq:federated_fragmentation}
\end{equation}
where $\mathcal{K}_{c}$ contains the clients observing class $c$.

\begin{figure}[t]
    \centering
    \includegraphics[width=\linewidth]{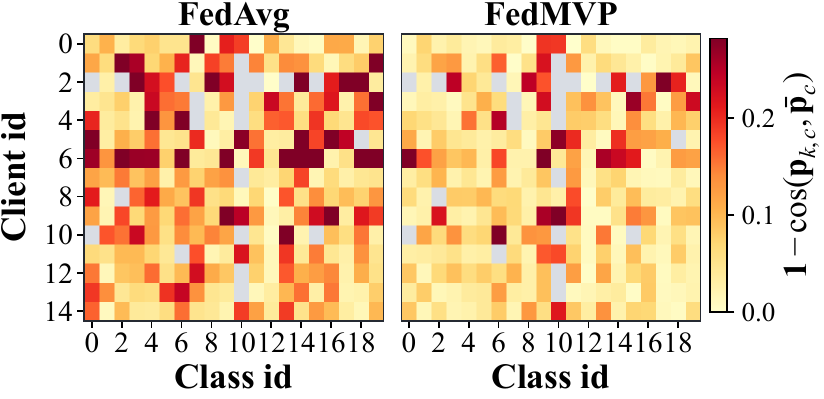}
    \caption{Client-to-global prototype discrepancies for FedAvg (left) and FedMVP (right) with 15 clients; darker cells indicate larger cosine distance.}
    \label{fig:federated_fragmentation}
\end{figure}

Fig.~\ref{fig:federated_fragmentation} reveals substantial class- and client-wise variation, with some categories retaining stable representations while others suffer severe forgetting or poorly calibrated OOD boundaries across clients. FedMVP narrows the overall spread relative to FedAvg, indicating that multimodal information helps reduce part of this inconsistency. However, considerable fragmentation remains, especially for classes exposed to heterogeneous local distributions. This suggests that multimodal fusion alone cannot ensure consistent category retention or robust OOD separation across clients, motivating explicit mechanisms for preserving and coordinating distributed category memories.

\section{Methodology}
\label{sec:methodology}

\subsection{Overview}
\label{sec:method_overview}

\begin{figure*}[t]
    \centering
    \includegraphics[width=\textwidth]{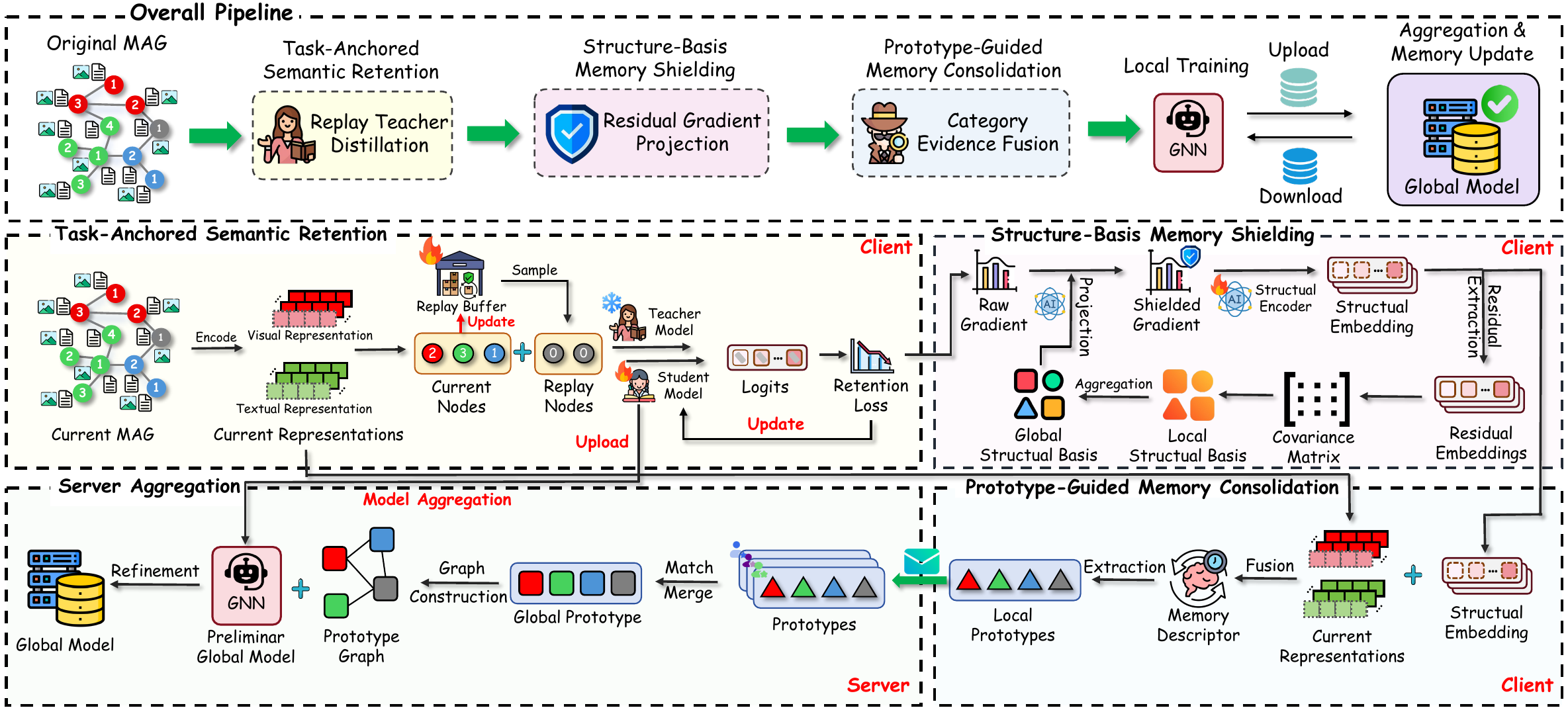}
    \caption{Overview of FedOGL. At each incremental stage, clients learn from private multimodal graph streams and exchange only model updates and compact memory summaries with the server. FedOGL integrates task-anchored semantic retention, structure-basis memory shielding, and prototype-guided memory consolidation to preserve historical semantic-structural knowledge while adapting to emerging classes and maintaining open-world rejection.}
    \label{fig:framework}
\end{figure*}

Fig.~\ref{fig:framework} illustrates the overall framework of \textbf{FedOGL}. At each incremental stage, clients learn from private multimodal graph streams, while the server aggregates model updates and compact memory summaries. We denote the semantic and structural representations used in later modules as $\mathbf{s}_{i}^{t}$ and $\mathbf{g}_{i}^{t}$, respectively. Following the three blocks in Fig.~\ref{fig:framework}, \emph{Task-Anchored Semantic Retention} combines replay with teacher distillation, \emph{Structure-Basis Memory Shielding} applies residual gradient projection using a global structural basis, and \emph{Prototype-Guided Memory Consolidation} fuses category evidence into local prototypes that are consolidated on the server. Together, they preserve historical semantic-structural memory while learning emerging classes.

\subsection{Task-Anchored Semantic Retention}
\label{sec:method_retention}

\textbf{Motivation.}
When new classes arrive, the semantic representation pathway can quickly adapt to current categories but may overwrite the modality-specific semantics learned for old ones. This is the main source of modality-semantic overwriting. To preserve historical decision behavior without directly transmitting raw data, each client anchors current learning to a task-start teacher and a small replay window.

\textbf{Replay and teacher distillation.}
At stage $t$, client $k$ freezes the received global model as a task-start teacher $\theta_{k,t}^{-}$, following established replay and distillation strategies~\cite{rebuffi2017icarl}. Let $\mathcal{D}_{k}^{t}$ denote current-stage training nodes and let $\mathcal{R}_{k}(t)=\{t-q,\ldots,t-1\}$ denote replayed historical stages, where $q=\min(M,t-1)$. FedOGL optimizes the task-anchored retention objective in Eq.~\eqref{eq:retention_loss}:
\begin{equation}
\begin{aligned}
\mathcal{L}_{k,t}^{\mathrm{ret}}
=&\;
\underbrace{\mathcal{L}_{\mathrm{ce}}(\mathcal{D}_{k}^{t};\theta)}
_{\text{new-class plasticity}}
+
\lambda_{\mathrm{rep}}
\underbrace{
\mathbb{E}_{r\in\mathcal{R}_{k}(t)}
\mathcal{L}_{\mathrm{ce}}(\mathcal{D}_{k}^{r};\theta)}
_{\text{historical replay}}
\\
&+
\lambda_{\mathrm{dis}}\tau^{2}
\underbrace{
\mathbb{E}_{r\in\mathcal{R}_{k}(t)}
\mathrm{KL}\!\left(
\sigma(\boldsymbol{\ell}_{\theta_{k,t}^{-}}^{r}/\tau)
\Vert
\sigma(\boldsymbol{\ell}_{\theta}^{r}/\tau)
\right)}
_{\text{task-start boundary anchoring}} .
\end{aligned}
\label{eq:retention_loss}
\end{equation}
The replay term preserves supervised information from recent historical stages, while task-start distillation constrains the model not to deviate sharply from the pre-update boundary.

\textbf{Open-world rejection.}
To prevent unknown samples from being forced into the current known class space, FedOGL uses maximum softmax probability as the known-class confidence in Eq.~\eqref{eq:ood_score}~\cite{hendrycks2017baseline}:
\begin{equation}
\begin{aligned}
\rho_i^t &= \max_{c\in\mathcal{C}_{\mathrm{K}}^t} p_{i,c}^{t},
\hat{y}_i =
\begin{cases}
\arg\max_{c\in\mathcal{C}_{\mathrm{K}}^t} p_{i,c}^{t}, & \rho_i^t\ge \delta_t,\\
\mathrm{OOD}, & \rho_i^t<\delta_t .
\end{cases}
\end{aligned}
\label{eq:ood_score}
\end{equation}
This confidence rule turns the retained known-class boundary into an open-world decision, while the next module protects the structural supports behind this boundary.

\subsection{Structure-Basis Memory Shielding}
\label{sec:method_basis}

\textbf{Motivation.}
In graph learning, old categories are not preserved only by classifier logits. They are also supported by topology-dependent propagation directions. If new-task gradients freely update the structural representation pathway, these historical graph-propagation directions can be erased even when label-level replay is used. FedOGL therefore protects structural memory through a global subspace basis.

\textbf{Client-side residual subspace extraction.}
After stage $t$, each participating client extracts structural embeddings $\mathbf{Z}_{k,t}^{\mathrm{str}}$ from training nodes. With centered embeddings $\mathbf{H}_{k,t}=\mathbf{Z}_{k,t}^{\mathrm{str}}-\mathbf{1}\boldsymbol{\mu}_{k,t}^{\top}$, the client removes directions already explained by $\mathbf{U}_{t-1}$ and computes the residual structural basis in Eq.~\eqref{eq:local_basis}:
\begin{equation}
\begin{aligned}
\widetilde{\mathbf{H}}_{k,t}
&=
\mathbf{H}_{k,t}
(\mathbf{I}-\mathbf{U}_{t-1}\mathbf{U}_{t-1}^{\top}),\\
\mathbf{C}_{k,t}
&=
\widetilde{\mathbf{H}}_{k,t}^{\top}
\mathbf{W}_{k,t}
\widetilde{\mathbf{H}}_{k,t},
\mathbf{U}_{k,t}
=
\mathrm{TopEig}_{r_l}(\mathbf{C}_{k,t}),
\end{aligned}
\label{eq:local_basis}
\end{equation}
where $\mathbf{W}_{k,t}$ optionally applies class-balanced sample weights. Each $\mathbf{U}_{k,t}$ is a compact structural summary rather than raw graph data.

\textbf{Server-side structural memory aggregation.}
The server aggregates local structural memory in Eq.~\eqref{eq:global_basis} by summing projectors instead of directly averaging basis vectors:
\begin{equation}
\begin{aligned}
\mathbf{S}_{t}
&=
\rho\,\mathbf{U}_{t-1}\mathbf{U}_{t-1}^{\top}
+
\sum_{k\in\mathcal{S}_{t}}
\alpha_k
\mathbf{U}_{k,t}\mathbf{U}_{k,t}^{\top},
\\
\mathbf{U}_{t}
&=
\mathrm{TopEig}_{r}(\mathbf{S}_{t}),
\alpha_k=\frac{n_{k,t}}{\sum_{j\in\mathcal{S}_{t}}n_{j,t}} .
\end{aligned}
\label{eq:global_basis}
\end{equation}
This projector-based aggregation preserves subspace geometry and yields a global basis that summarizes historical graph-propagation directions across clients.

\textbf{Residual gradient projection.}
The updated basis $\mathbf{U}_{t}$ is used to shield structural gradients in the next incremental stage through Eq.~\eqref{eq:gradient_shield}:
\begin{equation}
\begin{aligned}
\mathcal{P}_{\perp}^{t}(\mathbf{G})
&=
\mathbf{G}(\mathbf{I}-\mathbf{U}_{t}\mathbf{U}_{t}^{\top}),\\
\widetilde{\nabla}_{\Omega_{\mathrm{str}}}\mathcal{L}_{k,t}
&=
\mathcal{P}_{\perp}^{t}
\left(
\nabla_{\Omega_{\mathrm{str}}}\mathcal{L}_{k,t}
\right),
\end{aligned}
\label{eq:gradient_shield}
\end{equation}
where $\Omega_{\mathrm{str}}$ denotes structural parameters aligned with the protected basis. This shielding allows the model to learn new structural patterns while suppressing updates that would overwrite historical graph-propagation memory. The protected structural memory is then complemented by a server-side category memory.

\subsection{Prototype-Guided Memory Consolidation}
\label{sec:method_prototype}

\textbf{Motivation.}
In federated open-world graph streams, no client observes the complete class space or the complete semantic-structural support of old categories. Thus, old-class memory is fragmented across clients and cannot be recovered by parameter averaging alone. FedOGL addresses this issue by maintaining a compact server-side prototype memory that consolidates category knowledge across clients without directly transmitting raw graph data.

\textbf{Category evidence fusion and prototype construction.}
For each observed class $c$, client $k$ summarizes its local semantic-structural support into one or more prototypes. We first define the node-level memory descriptor in Eq.~\eqref{eq:node_descriptor}:
\begin{equation}
\mathbf{z}_{i}^{t}
=
\Pi_{\omega}
\left(
[\mathbf{s}_{i}^{t}\Vert\mathbf{g}_{i}^{t}\Vert\mathbf{a}_{i}^{t}]
\right),
\label{eq:node_descriptor}
\end{equation}
where $\Pi_{\omega}(\cdot)$ aligns semantic features, structural features, and modality-availability indicators into a shared prototype space. For each class mode $q$, the local prototype and its support size are computed by Eq.~\eqref{eq:local_proto}:
\begin{equation}
\mathbf{p}_{k,c,q}^{t}
=
\frac{1}{|\mathcal{V}_{k,c,q}^{t}|}
\sum_{i\in\mathcal{V}_{k,c,q}^{t}}\mathbf{z}_{i}^{t},
n_{k,c,q}^{t}
=
|\mathcal{V}_{k,c,q}^{t}|.
\label{eq:local_proto}
\end{equation}
Here, $q$ indexes optional within-class modes. These prototypes provide compact class-level summaries of local semantic-structural memory.

\textbf{Server-side memory update.}
The server maintains a bounded memory bank $\mathcal{M}_{c}=\{(\bar{\mathbf{p}}_{c,j},n_{c,j})\}_{j=1}^{S}$ for each class. Given an incoming prototype $\mathbf{p}$ with support size $n_p$, FedOGL first finds the closest memory slot and then performs the support-weighted consolidation in Eq.~\eqref{eq:proto_update}:
\begin{equation}
\begin{aligned}
j^{*}
&=
\arg\max_{j\le S}
\cos(\mathbf{p},\bar{\mathbf{p}}_{c,j}),\\
(\bar{\mathbf{p}}_{c,j^{*}},n_{c,j^{*}})
&\leftarrow
\left(
\frac{n_{c,j^{*}}\bar{\mathbf{p}}_{c,j^{*}}+n_p\mathbf{p}}
{n_{c,j^{*}}+n_p},
\;
n_{c,j^{*}}+n_p
\right),
\end{aligned}
\label{eq:proto_update}
\end{equation}
if $\cos(\mathbf{p},\bar{\mathbf{p}}_{c,j^{*}})>\eta$. Otherwise, the incoming prototype occupies an empty slot or replaces the weakest slot. This update accumulates cross-client category memory while keeping raw nodes, edges, and modalities local.

\textbf{Prototype-graph consolidation.}
To further consolidate the memory bank, the server constructs the lightweight prototype graph in Eq.~\eqref{eq:proto_graph}:
\begin{equation}
\mathcal{G}_{P}^{t}
=
(\mathcal{P}^{t},\mathcal{E}_{P}^{t}),
\mathcal{E}_{P}^{t}
=
\mathrm{KNN}_{\kappa}(\mathcal{P}^{t})
\cup
\mathcal{E}_{\mathrm{same}},
\label{eq:proto_graph}
\end{equation}
where $\mathcal{E}_{\mathrm{same}}$ connects prototypes with the same class label. The global model is refined on this compact memory by Eq.~\eqref{eq:proto_loss}:
\begin{equation}
\mathcal{L}_{\mathrm{mem}}
=
\gamma_t
\left[
\lambda_{\mathrm{lab}}
\mathcal{L}_{\mathrm{ce}}(\mathcal{G}_{P}^{t};\theta)
+
\lambda_{\mathrm{tr}}
\mathrm{KL}
\left(
\bar{\mathbf{q}}_{\mathrm{cli}}
\Vert
\mathbf{q}_{\theta}
\right)
\right],
\label{eq:proto_loss}
\end{equation}
where $\bar{\mathbf{q}}_{\mathrm{cli}}$ denotes trajectory-aware soft targets aggregated from recent client models. Thus, prototype memory serves as a compact federated category substrate, complementing the protected structural basis and reducing federated memory fragmentation in future stages.

\textbf{End-to-end update.}\ 
At each stage, a client encodes its current multimodal attributed graph into visual, textual, and structural representations. Current and replayed nodes are learned with a frozen task-start teacher, while the global structural basis projects raw structural gradients into shielded gradients. The resulting semantic and structural evidence is fused into memory descriptors, from which the client extracts local prototypes and a residual structural basis. The client uploads these summaries with its model update; the server aggregates model parameters and local bases, matches and merges the prototypes, constructs the prototype graph, and refines the preliminary global model. The updated global model and structural basis support the next round, matching the client--server flow in Fig.~\ref{fig:framework}.

\begin{table*}[!t]
    \centering
\scriptsize
    \setlength{\tabcolsep}{2.3pt}
    \renewcommand{\arraystretch}{1.05}
    \resizebox{0.99\textwidth}{!}{
    \begin{tabular}{llccccccccc}
    \specialrule{1.5pt}{1.5pt}{1.5pt}
    \multicolumn{2}{c}{\textbf{Category}} &
    \multicolumn{9}{c}{\textbf{Node Classification: AM$\uparrow$ / FM$\downarrow$ / FPR$\downarrow$ (\%)}} \\
    \cmidrule{1-11}
    \multicolumn{2}{c}{\textbf{Dataset}} &
    \multicolumn{3}{c}{\textbf{Grocery}} &
    \multicolumn{3}{c}{\textbf{RedditS}} &
    \multicolumn{3}{c}{\textbf{Ele-Fashion}} \\
    \cmidrule{1-11}
    & \textbf{Method} &
    \textbf{AM$\uparrow$} & \textbf{FM$\downarrow$} & \textbf{FPR$\downarrow$} &
    \textbf{AM$\uparrow$} & \textbf{FM$\downarrow$} & \textbf{FPR$\downarrow$} &
    \textbf{AM$\uparrow$} & \textbf{FM$\downarrow$} & \textbf{FPR$\downarrow$} \\
    \midrule
    \multirow{3}{*}{\textbf{FL}}
    & \textbf{FedAvg}
    & $59.94_{\scriptstyle \pm 0.48}$ & $10.83_{\scriptstyle \pm 0.32}$ & $58.55_{\scriptstyle \pm 0.76}$
    & $79.91_{\scriptstyle \pm 0.37}$ & $9.12_{\scriptstyle \pm 0.28}$ & $10.47_{\scriptstyle \pm 0.64}$
    & $55.94_{\scriptstyle \pm 0.84}$ & $9.41_{\scriptstyle \pm 0.37}$ & $74.10_{\scriptstyle \pm 1.71}$ \\
    & \textbf{FedMVP}
    & $64.75_{\scriptstyle \pm 0.44}$ & $11.46_{\scriptstyle \pm 0.34}$ & $50.98_{\scriptstyle \pm 0.71}$
    & $79.57_{\scriptstyle \pm 0.46}$ & $11.23_{\scriptstyle \pm 0.34}$ & $11.86_{\scriptstyle \pm 0.68}$
    & $58.03_{\scriptstyle \pm 0.64}$ & $10.39_{\scriptstyle \pm 0.44}$ & $67.36_{\scriptstyle \pm 1.96}$ \\
    \midrule
    \multirow{2}{*}{\textbf{FGL}}
    & \textbf{PRISM}
    & $59.99_{\scriptstyle \pm 3.16}$ & $12.56_{\scriptstyle \pm 0.92}$ & $60.79_{\scriptstyle \pm 1.71}$
    & $83.43_{\scriptstyle \pm 2.31}$ & $11.44_{\scriptstyle \pm 3.14}$ & $12.50_{\scriptstyle \pm 1.01}$
    & $53.63_{\scriptstyle \pm 1.27}$ & $11.19_{\scriptstyle \pm 1.17}$ & $70.43_{\scriptstyle \pm 1.66}$ \\
    & \textbf{FedSSP}
    & $55.88_{\scriptstyle \pm 0.49}$ & $12.35_{\scriptstyle \pm 0.37}$ & $57.75_{\scriptstyle \pm 0.62}$
    & $76.43_{\scriptstyle \pm 1.21}$ & $10.36_{\scriptstyle \pm 2.16}$ & $21.27_{\scriptstyle \pm 2.85}$
    & $56.66_{\scriptstyle \pm 0.48}$ & $9.44_{\scriptstyle \pm 0.32}$ & $66.59_{\scriptstyle \pm 1.92}$ \\
    & \textbf{FedIIH}
    & $47.50_{\scriptstyle \pm 0.47}$ & $8.23_{\scriptstyle \pm 0.25}$ & $67.03_{\scriptstyle \pm 0.49}$
    & \underline{$85.09_{\scriptstyle \pm 0.33}$} & $6.78_{\scriptstyle \pm 0.21}$ & $12.15_{\scriptstyle \pm 0.32}$
    & $66.58_{\scriptstyle \pm 0.49}$ & \underline{$5.95_{\scriptstyle \pm 0.25}$} & $56.56_{\scriptstyle \pm 1.33}$ \\
    \midrule
    \multirow{4}{*}{\textbf{Open-World}}
    & \textbf{POWER}
    & $\textbf{67.80}_{\scriptstyle \pm \textbf{2.08}}$ & \underline{$6.70_{\scriptstyle \pm 0.46}$} & \underline{$42.25_{\scriptstyle \pm 1.56}$}
    & $74.35_{\scriptstyle \pm 1.87}$ & $\textbf{3.49}_{\scriptstyle \pm \textbf{2.28}}$ & $22.55_{\scriptstyle \pm 0.35}$
    & $65.58_{\scriptstyle \pm 2.90}$ & $6.78_{\scriptstyle \pm 1.94}$ & \underline{$52.63_{\scriptstyle \pm 2.76}$} \\
    & \textbf{TopoOOD}
    & $62.59_{\scriptstyle \pm 3.16}$ & $11.58_{\scriptstyle \pm 2.66}$ & $56.81_{\scriptstyle \pm 3.29}$
    & $83.77_{\scriptstyle \pm 0.60}$ & $10.90_{\scriptstyle \pm 2.55}$ & \underline{$9.61_{\scriptstyle \pm 0.95}$}
    & \underline{$67.85_{\scriptstyle \pm 1.09}$} & $8.83_{\scriptstyle \pm 1.17}$ & $63.57_{\scriptstyle \pm 2.93}$ \\
    & \textbf{GRASP}
    & $57.72_{\scriptstyle \pm 1.86}$ & $10.46_{\scriptstyle \pm 2.14}$ & $59.25_{\scriptstyle \pm 3.02}$
    & $76.53_{\scriptstyle \pm 1.53}$ & $9.52_{\scriptstyle \pm 1.63}$ & $12.39_{\scriptstyle \pm 0.95}$
    & $60.18_{\scriptstyle \pm 1.07}$ & $9.00_{\scriptstyle \pm 2.61}$ & $61.12_{\scriptstyle \pm 2.36}$ \\
    & \textbf{CLIPN}
    & $47.62_{\scriptstyle \pm 0.51}$ & $9.87_{\scriptstyle \pm 0.29}$ & $70.84_{\scriptstyle \pm 0.70}$
    & $84.39_{\scriptstyle \pm 0.37}$ & $9.45_{\scriptstyle \pm 0.28}$ & $10.18_{\scriptstyle \pm 0.05}$
    & $62.37_{\scriptstyle \pm 0.60}$ & $12.04_{\scriptstyle \pm 0.76}$ & $68.74_{\scriptstyle \pm 0.99}$ \\
    \midrule
    \multirow{1}{*}{\textbf{Ours}}
    & \textbf{FedOGL}
    & \underline{$67.24_{\scriptstyle \pm 0.80}$} & $\textbf{3.19}_{\scriptstyle \pm \textbf{0.05}}$ & $\textbf{31.61}_{\scriptstyle \pm \textbf{0.67}}$
    & $\textbf{92.35}_{\scriptstyle \pm \textbf{0.34}}$ & \underline{$5.64_{\scriptstyle \pm 0.47}$} & $\textbf{8.04}_{\scriptstyle \pm \textbf{0.03}}$
    & $\textbf{71.11}_{\scriptstyle \pm \textbf{0.83}}$ & $\textbf{4.70}_{\scriptstyle \pm \textbf{0.38}}$ & $\textbf{45.97}_{\scriptstyle \pm \textbf{1.88}}$ \\
    \specialrule{1.3pt}{2.0pt}{1.0pt}
    \end{tabular}}
    \vspace{1.5mm}
\scriptsize
    \setlength{\tabcolsep}{2.3pt}
    \renewcommand{\arraystretch}{1.05}
    \resizebox{0.99\textwidth}{!}{
    \begin{tabular}{llccccccccc}
    \specialrule{1.5pt}{1.5pt}{1.5pt}
    \multicolumn{2}{c}{\textbf{Category}} &
    \multicolumn{9}{c}{\textbf{Modality Retrieval: AM$\uparrow$ / FM$\downarrow$ / FPR$\downarrow$ (\%)}} \\
    \cmidrule{1-11}
    \multicolumn{2}{c}{\textbf{Dataset}} &
    \multicolumn{3}{c}{\textbf{KU}} &
    \multicolumn{3}{c}{\textbf{QB}} &
    \multicolumn{3}{c}{\textbf{Bili-Cartoon}} \\
    \cmidrule{1-11}
    & \textbf{Method} &
    \textbf{AM$\uparrow$} & \textbf{FM$\downarrow$} & \textbf{FPR$\downarrow$} &
    \textbf{AM$\uparrow$} & \textbf{FM$\downarrow$} & \textbf{FPR$\downarrow$} &
    \textbf{AM$\uparrow$} & \textbf{FM$\downarrow$} & \textbf{FPR$\downarrow$} \\
    \midrule
    \multirow{3}{*}{\textbf{FL}}
    & \textbf{FedAvg}
    & $43.46_{\scriptstyle \pm 0.42}$ & $14.23_{\scriptstyle \pm 0.43}$ & $79.08_{\scriptstyle \pm 0.84}$
    & $60.73_{\scriptstyle \pm 0.47}$ & $11.67_{\scriptstyle \pm 0.35}$ & $67.45_{\scriptstyle \pm 0.71}$
    & $58.88_{\scriptstyle \pm 0.48}$ & $11.12_{\scriptstyle \pm 0.33}$ & $59.42_{\scriptstyle \pm 0.67}$ \\
    & \textbf{FedMVP}
    & $43.93_{\scriptstyle \pm 0.39}$ & $15.56_{\scriptstyle \pm 0.47}$ & $78.75_{\scriptstyle \pm 0.81}$
    & $70.28_{\scriptstyle \pm 0.43}$ & $13.23_{\scriptstyle \pm 0.40}$ & $47.03_{\scriptstyle \pm 0.75}$
    & $54.35_{\scriptstyle \pm 0.36}$ & $14.87_{\scriptstyle \pm 0.45}$ & $63.49_{\scriptstyle \pm 0.89}$ \\
    \midrule
    \multirow{2}{*}{\textbf{FGL}}
    & \textbf{PRISM}
    & $42.58_{\scriptstyle \pm 3.41}$ & $16.62_{\scriptstyle \pm 3.01}$ & $76.44_{\scriptstyle \pm 2.42}$
    & $69.99_{\scriptstyle \pm 1.56}$ & $16.75_{\scriptstyle \pm 2.60}$ & $49.74_{\scriptstyle \pm 1.11}$
    & $56.71_{\scriptstyle \pm 1.94}$ & $16.19_{\scriptstyle \pm 3.04}$ & $59.21_{\scriptstyle \pm 1.91}$ \\
    & \textbf{FedSSP}
    & $44.53_{\scriptstyle \pm 0.38}$ & $13.34_{\scriptstyle \pm 0.40}$ & $78.56_{\scriptstyle \pm 0.81}$
    & $65.50_{\scriptstyle \pm 0.43}$ & $10.92_{\scriptstyle \pm 0.33}$ & $48.56_{\scriptstyle \pm 0.58}$
    & $57.99_{\scriptstyle \pm 0.44}$ & $11.34_{\scriptstyle \pm 0.34}$ & $58.23_{\scriptstyle \pm 0.61}$ \\
    & \textbf{FedIIH}
    & $42.49_{\scriptstyle \pm 0.35}$ & $18.76_{\scriptstyle \pm 0.26}$ & $79.78_{\scriptstyle \pm 0.83}$
    & $\textbf{78.45}_{\scriptstyle \pm \textbf{0.37}}$ & \underline{$7.89_{\scriptstyle \pm 0.24}$} & {$28.67_{\scriptstyle \pm 0.48}$}
    & $57.29_{\scriptstyle \pm 0.41}$ & \underline{$7.23_{\scriptstyle \pm 0.22}$} & $47.89_{\scriptstyle \pm 0.52}$ \\
    \midrule
    \multirow{4}{*}{\textbf{Open-World}}
    & \textbf{POWER}
    & $41.31_{\scriptstyle \pm 0.58}$ & $13.21_{\scriptstyle \pm 3.40}$ & $75.11_{\scriptstyle \pm 2.39}$
    & $72.73_{\scriptstyle \pm 2.36}$ & $14.50_{\scriptstyle \pm 2.46}$ & $35.52_{\scriptstyle \pm 3.08}$
    & $60.19_{\scriptstyle \pm 0.98}$ & $15.32_{\scriptstyle \pm 2.47}$ & $47.25_{\scriptstyle \pm 1.44}$ \\
    & \textbf{TopoOOD}
    & $41.99_{\scriptstyle \pm 2.15}$ & $16.04_{\scriptstyle \pm 3.04}$ & $\textbf{40.67}_{\scriptstyle \pm \textbf{3.12}}$
    & $58.52_{\scriptstyle \pm 2.43}$ & $12.50_{\scriptstyle \pm 1.70}$ & $62.81_{\scriptstyle \pm 1.52}$
    & $56.24_{\scriptstyle \pm 1.17}$ & $13.40_{\scriptstyle \pm 1.92}$ & $52.12_{\scriptstyle \pm 2.84}$ \\
    & \textbf{GRASP}
    & \underline{$44.80_{\scriptstyle \pm 3.40}$} & $14.77_{\scriptstyle \pm 1.51}$ & $69.35_{\scriptstyle \pm 1.46}$
    & $62.60_{\scriptstyle \pm 3.02}$ & $12.06_{\scriptstyle \pm 0.21}$ & $58.91_{\scriptstyle \pm 0.73}$
    & \underline{$60.65_{\scriptstyle \pm 1.23}$} & $11.57_{\scriptstyle \pm 0.91}$ & \underline{$47.15_{\scriptstyle \pm 0.90}$} \\
    & \textbf{CLIPN}
    & $43.33_{\scriptstyle \pm 0.38}$ & \underline{$12.89_{\scriptstyle \pm 0.39}$} & $78.67_{\scriptstyle \pm 0.80}$
    & $73.18_{\scriptstyle \pm 0.40}$ & $11.15_{\scriptstyle \pm 0.33}$ & \underline{$28.65_{\scriptstyle \pm 0.76}$}
    & $53.02_{\scriptstyle \pm 0.35}$ & $10.56_{\scriptstyle \pm 0.31}$ & $53.78_{\scriptstyle \pm 0.89}$ \\
    \midrule
    \multirow{1}{*}{\textbf{Ours}}
    & \textbf{FedOGL}
    & $\textbf{45.85}_{\scriptstyle \pm \textbf{0.70}}$ & $\textbf{10.38}_{\scriptstyle \pm \textbf{0.59}}$ & \underline{$51.76_{\scriptstyle \pm 0.60}$}
    & \underline{$76.35_{\scriptstyle \pm 0.48}$} & $\textbf{3.88}_{\scriptstyle \pm \textbf{0.25}}$ & $\textbf{28.60}_{\scriptstyle \pm \textbf{0.51}}$
    & $\textbf{65.02}_{\scriptstyle \pm \textbf{0.50}}$ & $\textbf{3.65}_{\scriptstyle \pm \textbf{0.22}}$ & $\textbf{31.40}_{\scriptstyle \pm \textbf{0.53}}$ \\
    \specialrule{1.3pt}{2.0pt}{1.0pt}
    \end{tabular}}
    \caption{Federated open-world continual graph-learning performance on node classification and modality retrieval. Results are mean$\pm$std (\%). Best results are shown in bold and second-best results are underlined.}
    \label{tab:fedogl_overall_node}
\end{table*}

\section{Experiments}
\label{sec:fedogl_exp}

We evaluate \textbf{FedOGL} through 5 questions. \textbf{Q1: Overall effectiveness.} Does FedOGL jointly improve final task performance (AM), preserve previously learned tasks (FM), and reject unknown samples (FPR$_{95}$) across node classification and modality retrieval? \textbf{Q2: Retention dynamics.} How well does FedOGL retain early-task knowledge throughout class-incremental evolution, rather than only at the final stage? \textbf{Q3: Component effectiveness.} How much do replay, distillation, structural projection, and prototype transfer contribute to the complete framework? \textbf{Q4: Robustness and optimization.} How does FedOGL behave under different communication rounds, federation scales, and hyperparameter settings? \textbf{Q5: Practical efficiency.} What communication and runtime overhead does FedOGL introduce relative to the baselines?

\subsection{Experimental Setup}
\label{sec:fedogl_setup}

\textbf{Baselines and Metrics.}\ We compare FedOGL with 3 categories of baselines: FL methods (FedAvg~\cite{mcmahan2017fedavg}, PRISM~\cite{prism2026}, and FedMVP~\cite{fedmvp2023}); FGL methods (FedSSP~\cite{fedssp2024} and FedIIH~\cite{fediih2025}); and open-world methods covering continual retention and OOD detection (POWER~\cite{zhu2024federated}, TopoOOD~\cite{li2024topoood}, GRASP~\cite{liu2023grasp}, and CLIPN~\cite{clipn2023}). As existing baselines rarely cover the full setting, we apply minimal
adaptations: centralized methods use local training with FedAvg
aggregation, while non-continual methods use a shared class-balanced
replay budget. Implementation details are provided in Section~2 of the
supporting material.

We report AM, FM, and FPR$_{95}$. AM denotes the mean performance
over all seen tasks after the final incremental stage, measured by
accuracy for node classification and Recall@1 for modality retrieval.
FM measures the average drop in each old task's corresponding metric
from its best historical performance to its final performance.
FPR$_{95}$ is the fraction of unknown samples accepted at 95\% TPR
on known samples. All metrics are reported as percentages, with higher
AM and lower FM/FPR$_{95}$ indicating better performance.

\textbf{Datasets and Protocol.}\ 
We use Grocery~\cite{ni2019amazon}, RedditS, and Ele-Fashion for node classification, and KU, QB, and Bili-Cartoon for cross-modal retrieval, following MM-OpenFGL~\cite{mmopenfgl2026} and MM-Graph~\cite{zhu2025mmgraph}. Louvain partitioning produces five non-IID client subgraphs~\cite{blondel2008louvain}; labels are then organized into class-incremental streams. Unless stated otherwise, every client participates in each round. Data construction and retrieval-pair generation are detailed in Section~1 of the supporting material. We report mean$\pm$std over 5 seeds.

\subsection{Overall Performance (Q1)}
\label{sec:fedogl_q1}

To answer \textbf{Q1}, we compare FedOGL with representative FL, FGL, and open-world baselines on node classification and cross-modal retrieval across all six datasets.

\textbf{Overall Results.}\ 
Table~\ref{tab:fedogl_overall_node} shows that FedOGL achieves the best or second-best performance for every metric across all datasets, with its most pronounced advantages in mitigating catastrophic forgetting and rejecting unknown classes. Specifically, FedOGL achieves the lowest FM on 5 datasets, with POWER attaining the lowest FM on RedditS. Similarly, it achieves the lowest FPR$_{95}$ on 5 datasets, with TopoOOD attaining the lowest FPR$_{95}$ on KU. Meanwhile, FedOGL maintains strong downstream task performance, achieving the best or second-best AM across all datasets. Compared with the strongest baseline for each metric, FedOGL reduces FM by $42.67\%$ (FedIIH) and FPR$_{95}$ by $28.31\%$ (POWER), while improving AM by $9.41\%$ (POWER).

\subsection{Anti-forgetting Analysis (Q2)}
\label{sec:fedogl_q2}

To answer \textbf{Q2}, we record task-wise accuracy after each task transition and track the average accuracy over all seen tasks for FedOGL and selected baselines.

\begin{figure}[!t]
    \centering
    \includegraphics[width=0.48\linewidth]{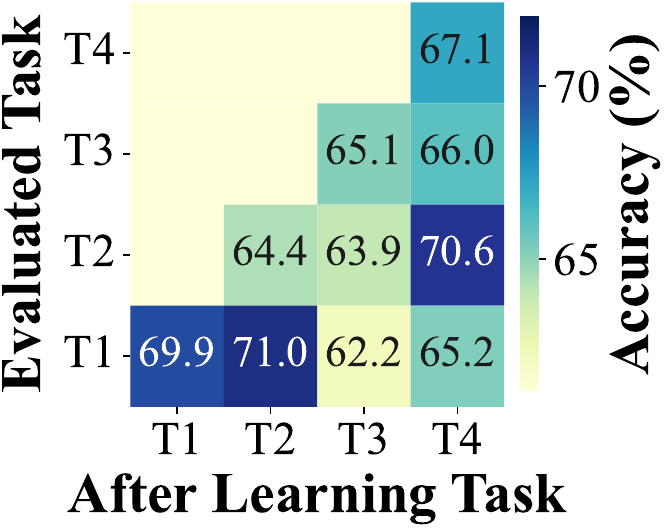}\hfill
    \includegraphics[width=0.48\linewidth]{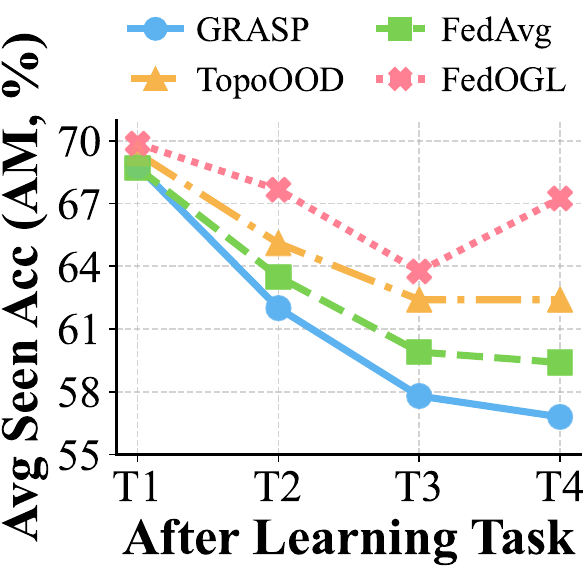}
    \caption{Anti-forgetting analysis on Grocery: task-wise retention (left) and average seen-task accuracy after each transition (right).}
    \label{fig:fedogl_forgetting}
\end{figure}

\textbf{Historical Task Retention.}\ 
Figure~\ref{fig:fedogl_forgetting} reports the retention dynamics on Grocery. The task-wise retention heatmap shows that FedOGL maintains stable performance on historical tasks as new classes are introduced. After learning all four tasks, the accuracies of the individual tasks remain within $65.2\%$--$70.6\%$. Consistently, the average-seen-task accuracy remains above those of the selected baselines after every transition. These results demonstrate that FedOGL mitigates forgetting throughout the entire incremental trajectory.

\subsection{Component Analysis (Q3)}
\label{sec:fedogl_q3}

To answer \textbf{Q3}, we conduct component ablations to quantify the contribution of each memory mechanism.

\begin{table}[t]
    \centering
    \small
    \setlength{\tabcolsep}{4.1pt}
    \renewcommand{\arraystretch}{1.08}
    \resizebox{\columnwidth}{!}{
    \begin{tabular}{lcccc}
    \toprule
    \textbf{Variant} &
    \multicolumn{2}{c}{\textbf{Grocery}} &
    \multicolumn{2}{c}{\textbf{Bili-Cartoon}} \\
    \cmidrule(lr){2-3}\cmidrule(lr){4-5}
    & \textbf{AM$\uparrow$} & \textbf{FM$\downarrow$}
    & \textbf{AM$\uparrow$} & \textbf{FM$\downarrow$} \\
    \midrule
    w/o Replay & $58.88_{\scriptstyle \pm 0.63}$ & $7.44_{\scriptstyle \pm 0.31}$ & $54.34_{\scriptstyle \pm 0.62}$ & $9.55_{\scriptstyle \pm 0.36}$ \\
    w/o Distillation & $61.73_{\scriptstyle \pm 0.55}$ & $5.61_{\scriptstyle \pm 0.27}$ & $58.40_{\scriptstyle \pm 0.56}$ & $7.21_{\scriptstyle \pm 0.30}$ \\
    w/o Structural Projection & $60.47_{\scriptstyle \pm 0.57}$ & $6.33_{\scriptstyle \pm 0.29}$ & $57.63_{\scriptstyle \pm 0.58}$ & $7.81_{\scriptstyle \pm 0.32}$ \\
    w/o Prototype Transfer & \underline{$62.41_{\scriptstyle \pm 0.49}$} & \underline{$4.31_{\scriptstyle \pm 0.21}$} & \underline{$60.18_{\scriptstyle \pm 0.51}$} & \underline{$6.16_{\scriptstyle \pm 0.25}$} \\
    \midrule
    Full FedOGL & $\textbf{67.24}_{\scriptstyle \pm \textbf{0.80}}$ & $\textbf{3.19}_{\scriptstyle \pm \textbf{0.05}}$ & $\textbf{65.02}_{\scriptstyle \pm \textbf{0.50}}$ & $\textbf{3.65}_{\scriptstyle \pm \textbf{0.22}}$ \\
    \bottomrule
    \end{tabular}}
    \caption{Ablation results on Grocery and Bili-Cartoon (mean$\pm$std, \%). Best results are bold and second-best results are underlined.}
    \vspace{-15pt}
    \label{tab:fedogl_ablation}
\end{table}

\textbf{Component Ablation.}
Table~\ref{tab:fedogl_ablation} shows that removing any component degrades performance. Replay has the largest impact on FM, while distillation, structural projection, and prototype transfer provide complementary benefits, confirming that all components contribute to FedOGL's overall performance.

\subsection{Robustness Analysis (Q4)}
\label{sec:fedogl_q4}

To answer \textbf{Q4}, we evaluate FedOGL under different communication rounds, federation scales, and hyperparameter settings to assess its optimization behavior and robustness.

\begin{figure}[t]
    \centering
    \includegraphics[width=\linewidth]{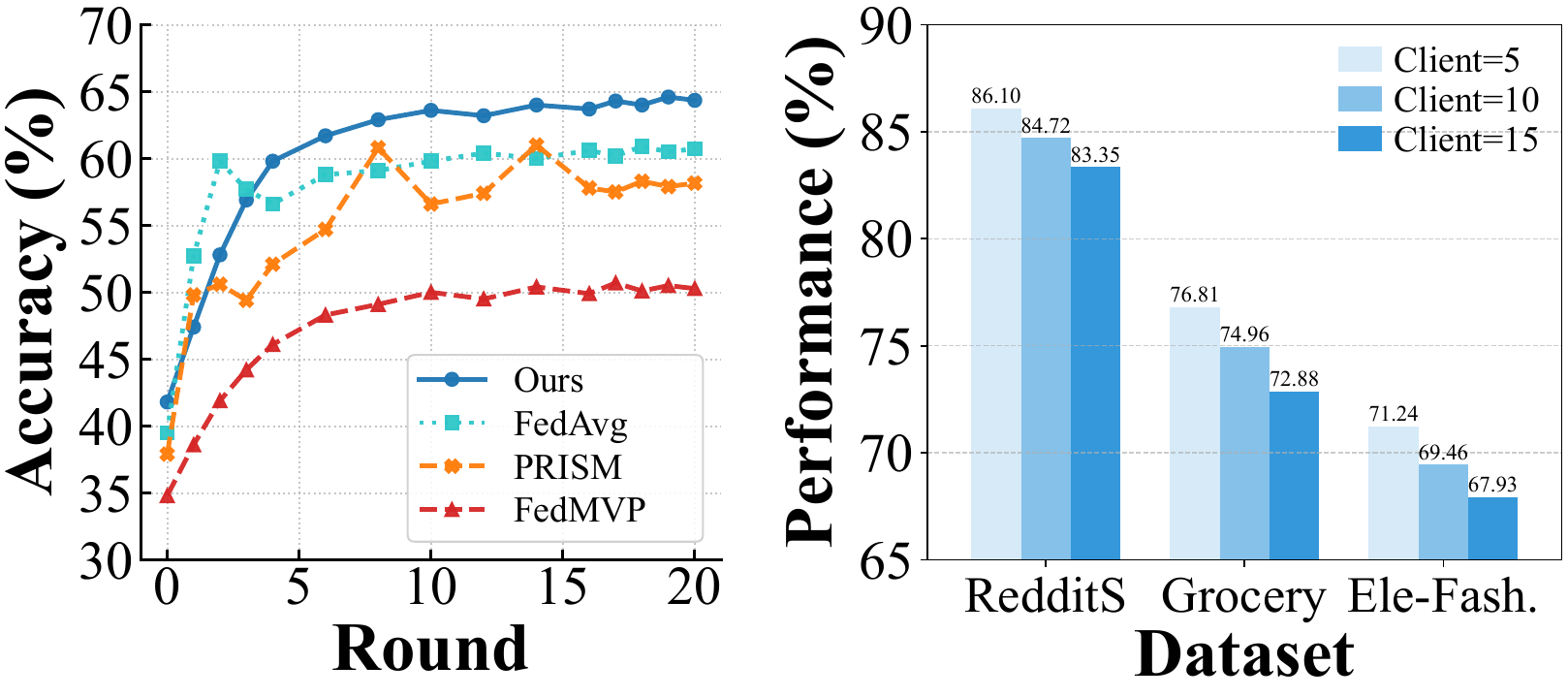}
    \caption{Federation-scale analysis. The two panels show QB retrieval convergence and client-scale robustness, respectively.}
    \label{fig:fedogl_clients}
\end{figure}

\textbf{Federation Scale.}\
Figure~\ref{fig:fedogl_clients} presents the convergence behavior and scalability results. On QB dataset, FedOGL converges to a higher retrieval performance than baselines. As the number of clients increases from 5 to 15, performance declines gradually across all three classification datasets, demonstrating the robustness of FedOGL under increasingly fragmented federations.

\begin{figure}[t]
    \centering
    \includegraphics[width=\linewidth]{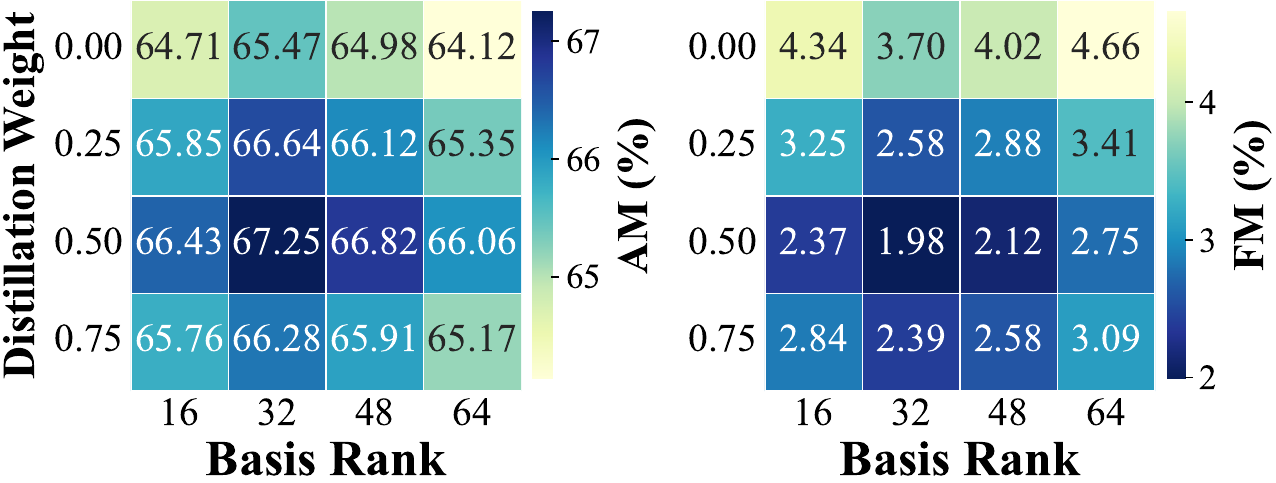}
    \caption{AM and FM sensitivity to the structure-basis rank and distillation weight.}
    \label{fig:fedogl_analysis}
    \vspace{-10pt}
\end{figure}

\textbf{Hyper-parameter Sensitivity.}\ 
Figure~\ref{fig:fedogl_analysis} studies the basis rank and distillation weight using both AM and FM. AM peaks and FM reaches its minimum near the default configuration, while nearby settings remain competitive. Very small or large values weaken the balance between adapting to new classes and preserving old ones, but there is no isolated knife-edge optimum. In practical terms, FedOGL remains stable across a reasonable neighborhood of settings and does not require unusually precise tuning to work well.

\subsection{Efficiency Analysis (Q5)}
\label{sec:fedogl_q5}

To answer \textbf{Q5}, we compare the communication cost and end-to-end runtime of FedOGL with representative baselines.

\textbf{Communication Efficiency.}
For node classification on Grocery and modality retrieval on QB, FedOGL incurs moderate communication cost: higher than the lightweight baselines but lower than POWER. Thus, its gains are not attributable to the largest communication budget. Detailed results are provided in Section~3 of the supporting material.

\textbf{Runtime Efficiency.}
In the available implementation logs, FedOGL runs close to FedAvg and substantially faster than the most expensive open-world baselines, although FedMVP remains faster. Overall, its additional memory operations introduce practical overhead. Detailed results and measurement caveats are provided in Section~4 of the supporting material.

\section{Conclusion}
\label{sec:conclusion}

We study federated open-world multimodal graph learning under class-incremental evolution. FedOGL combines semantic retention, structure-basis shielding, and prototype-guided consolidation to preserve distributed historical memory while learning new classes. Results on node classification and cross-modal retrieval show strong accuracy, retention, and unknown rejection with moderate communication overhead.

\bibliography{fedogl_refs}

@inproceedings{mcmahan2017fedavg,
  title={Communication-efficient learning of deep networks from decentralized data},
  author={McMahan, Brendan and Moore, Eider and Ramage, Daniel and Hampson, Seth and y Arcas, Blaise Aguera},
  booktitle={Artificial intelligence and statistics},
  pages={1273--1282},
  year={2017},
  organization={Pmlr}
}

@inproceedings{he2021fedgraphnn,
  title={Fedgraphnn: A federated learning benchmark system for graph neural networks},
  author={He, Chaoyang and Balasubramanian, Keshav and Ceyani, Emir and Yang, Carl and Xie, Han and Sun, Lichao and He, Lifang and Yang, Liangwei and Philip, S Yu and Rong, Yu and others},
  booktitle={ICLR 2021 Workshop on Distributed and Private Machine Learning (DPML)},
  year={2021}
}

@article{li2025openfgl,
  title={Openfgl: A comprehensive benchmark for federated graph learning},
  author={Li, Xunkai and Zhu, Yinlin and Pang, Boyang and Yan, Guochen and Yan, Yeyu and Li, Zening and Wu, Zhengyu and Zhang, Wentao and Li, Rong-Hua and Wang, Guoren},
  journal={arXiv preprint arXiv:2408.16288},
  year={2024}
}

@inproceedings{fedssp2024,
  title={FedSSP: Federated Graph Learning with Spectral Knowledge and Personalized Preference},
  author={Tan, Zihan and Wan, Guancheng and Huang, Wenke and Ye, Mang},
  booktitle={Advances in Neural Information Processing Systems},
  volume={37},
  year={2024}
}

@article{fediih2025,
  title={Modeling Inter-Intra Heterogeneity for Graph Federated Learning},
  author={Yu, Wentao and Chen, Shuo and Tong, Yongxin and Gu, Tianlong and Gong, Chen},
  journal={Proceedings of the AAAI Conference on Artificial Intelligence},
  volume={39},
  number={21},
  pages={22236--22244},
  year={2025},
  doi={10.1609/aaai.v39i21.34378}
}

@article{stage2026,
  title={STAGE: Tackling Semantic Drift in Multimodal Federated Graph Learning},
  author={Chen, Zekai and Wu, Xun and Li, Xunkai and Sun, Yihan and Li, Rong-Hua and Wang, Guoren},
  journal={arXiv preprint arXiv:2605.11919},
  year={2026}
}

@article{prism2026,
  title={PRISM: Topology-Aware Cross-Modal Imputation for Modality-Deficient Federated Graph Learning},
  author={Chen, Zekai and Zhang, Miao and Xing, Jiayang and Li, Xunkai and Wu, Xun and Li, Rong-Hua and Wang, Guoren},
  journal={arXiv preprint arXiv:2606.09301},
  year={2026}
}

@inproceedings{ssrm2023,
  title={Towards Robust Graph Incremental Learning on Evolving Graphs},
  author={Su, Junwei and Zou, Difan and Zhang, Zijun and Wu, Chuan},
  booktitle={Proceedings of the 40th International Conference on Machine Learning},
  series={Proceedings of Machine Learning Research},
  volume={202},
  pages={32728--32748},
  year={2023}
}

@inproceedings{rebuffi2017icarl,
  title={icarl: Incremental classifier and representation learning},
  author={Rebuffi, Sylvestre-Alvise and Kolesnikov, Alexander and Sperl, Georg and Lampert, Christoph H},
  booktitle={Proceedings of the IEEE conference on Computer Vision and Pattern Recognition},
  pages={2001--2010},
  year={2017}
}

@inproceedings{hendrycks2017baseline,
  title={A Baseline for Detecting Misclassified and Out-of-Distribution Examples in Neural Networks},
  author={Hendrycks, Dan and Gimpel, Kevin},
  booktitle={International Conference on Learning Representations},
  year={2017}
}

@inproceedings{zhu2024federated,
  title={Federated continual graph learning},
  author={Zhu, Yinlin and Hu, Miao and Wu, Di},
  booktitle={Proceedings of the 31st ACM SIGKDD Conference on Knowledge Discovery and Data Mining V. 2},
  pages={4203--4213},
  year={2025}
}

@inproceedings{fedmvp2023,
  title={FedMVP: Federated Multimodal Visual Prompt Tuning for Vision-Language Models},
  author={Singha, Mainak and Roy, Subhankar and Mehrotra, Sarthak and Jha, Ankit and Abdar, Moloud and Banerjee, Biplab and Ricci, Elisa},
  booktitle={Proceedings of the IEEE/CVF International Conference on Computer Vision},
  pages={17869--17878},
  year={2025}
}

@inproceedings{li2024topoood,
  title={Graph out-of-distribution detection goes neighborhood shaping},
  author={Bao, Tianyi and Wu, Qitian and Jiang, Zetian and Chen, Yiting and Sun, Jiawei and Yan, Junchi},
  booktitle={Forty-first International Conference on Machine Learning},
  year={2024}
}

@article{liu2023grasp,
  title={Revisiting score propagation in graph out-of-distribution detection},
  author={Ma, Longfei and Sun, Yiyou and Ding, Kaize and Liu, Zemin and Wu, Fei},
  journal={Advances in Neural Information Processing Systems},
  volume={37},
  pages={4341--4373},
  year={2024}
}

@inproceedings{clipn2023,
  title={Clipn for zero-shot ood detection: Teaching clip to say no},
  author={Wang, Hualiang and Li, Yi and Yao, Huifeng and Li, Xiaomeng},
  booktitle={Proceedings of the IEEE/CVF International Conference on Computer Vision},
  pages={1802--1812},
  year={2023}
}

@article{mmopenfgl2026,
  title={MM-OpenFGL: A Comprehensive Benchmark for Multimodal Federated Graph Learning},
  author={Li, Xunkai and Ai, Yuming and Zhu, Yinlin and Lu, Haodong and Zhang, Yi and Fu, Guohao and Fan, Bowen and Dai, Qiangqiang and Li, Rong-Hua and Wang, Guoren},
  journal={arXiv preprint arXiv:2601.22416},
  year={2026}
}

@inproceedings{zhu2025mmgraph,
  title={Mosaic of modalities: A comprehensive benchmark for multimodal graph learning},
  author={Zhu, Jing and Zhou, Yuhang and Qian, Shengyi and He, Zhongmou and Zhao, Tong and Shah, Neil and Koutra, Danai},
  booktitle={Proceedings of the IEEE/CVF Conference on Computer Vision and Pattern Recognition},
  pages={14215--14224},
  year={2025}
}

@inproceedings{ni2019amazon,
  title={Justifying recommendations using distantly-labeled reviews and fine-grained aspects},
  author={Ni, Jianmo and Li, Jiacheng and McAuley, Julian},
  booktitle={Proceedings of the 2019 conference on empirical methods in natural language processing and the 9th international joint conference on natural language processing (EMNLP-IJCNLP)},
  pages={188--197},
  year={2019}
}

@inproceedings{wang2023fashionklip,
  title={{FashionKLIP}: Enhancing E-Commerce Image-Text Retrieval with Fashion Multi-Modal Conceptual Knowledge Graph},
  author={Wang, Xiaodan and Wang, Chengyu and Li, Lei and Li, Zhixu and Chen, Ben and Jin, Linbo and Huang, Jun and Xiao, Yanghua and Gao, Ming},
  booktitle={Proceedings of the 61st Annual Meeting of the Association for Computational Linguistics (Volume 5: Industry Track)},
  pages={149--158},
  year={2023},
  publisher={Association for Computational Linguistics},
  doi={10.18653/v1/2023.acl-industry.16}
}

@inproceedings{wei2019mmgcn,
  title={{MMGCN}: Multi-modal Graph Convolution Network for Personalized Recommendation of Micro-video},
  author={Wei, Yinwei and Wang, Xiang and Nie, Liqiang and He, Xiangnan and Hong, Richang and Chua, Tat-Seng},
  booktitle={Proceedings of the 27th ACM International Conference on Multimedia},
  pages={1437--1445},
  year={2019},
  publisher={Association for Computing Machinery},
  doi={10.1145/3343031.3351034}
}

@article{blondel2008louvain,
  title={Fast unfolding of communities in large networks},
  author={Blondel, Vincent D and Guillaume, Jean-Loup and Lambiotte, Renaud and Lefebvre, Etienne},
  journal={Journal of statistical mechanics: theory and experiment},
  volume={2008},
  number={10},
  pages={P10008},
  year={2008}
}

\end{document}